\documentclass{article} 
\usepackage{iclr2022_conference,times}
\iclrfinalcopy
\pdfoutput=1

\usepackage{amsmath,amsfonts,bm}

\def\eqref#1{equation~\ref{#1}}

\def\1{\bm{1}}

\DeclareMathAlphabet{\mathsfit}{\encodingdefault}{\sfdefault}{m}{sl}
\SetMathAlphabet{\mathsfit}{bold}{\encodingdefault}{\sfdefault}{bx}{n}

\usepackage{hyperref}
\usepackage{url}

\usepackage{amsmath, amssymb}
\usepackage{multirow}
\usepackage{algorithmic}
\usepackage{algorithm}
\usepackage{graphbox}

\title{Adversarial Attack by Limited Point Cloud\\ Surface Modifications}

\author{Atrin Arya\thanks{Equal contribution.},$\ $  $\mbox{Hanieh Naderi}^*$ \& Shohreh Kasaei  \\
Department of Computer Engineering, Sharif University of Technology, Tehran, Iran\\
\texttt{\{aarya, hnaderi\}@ce.sharif.edu, kasaei@sharif.edu} 
}

\begin{document}

\maketitle

\begin{abstract}
Recent research has revealed that the security of deep neural networks that directly process 3D point clouds to classify objects can be threatened by adversarial samples. Although existing adversarial attack methods achieve high success rates, they do not restrict the point modifications enough to preserve the point cloud appearance. To overcome this shortcoming, two constraints are proposed. These include applying hard boundary constraints on the number of modified points and on the point perturbation norms. Due to the restrictive nature of the problem, the search space contains many local maxima. The proposed method addresses this issue by using a high step-size at the beginning of the algorithm to search the main surface of the point cloud fast and effectively. Then, in order to converge to the desired output, the step-size is gradually decreased. To evaluate the performance of the proposed method, it is run on the ModelNet40 and ScanObjectNN datasets by employing the state-of-the-art point cloud classification models; including PointNet, PointNet++, and DGCNN. The obtained results show that it can perform successful attacks and achieve state-of-the-art results by only a limited number of point modifications while preserving the appearance of the point cloud. Moreover, due to the effective search algorithm, it can perform successful attacks in just a few steps. Additionally, the proposed step-size scheduling algorithm shows an improvement of up to $14.5\%$ when adopted by other methods as well. The proposed method also performs effectively against popular defense methods.
\end{abstract}

\section{Introduction}
Deep Neural Networks (DNNs) have achieved considerable accuracy in various tasks (especially in classification). However, several recent research has shown that DNNs are not reliable enough \cite{carlini2017towards, goodfellow2015explaining, madry2019deep, naderi2021generating, brown2018unrestricted}. Adding small perturbations in the input data can cause the DNN to misclassify them with high confidence \cite{szegedy2014intriguing}. This new input data is called an adversarial sample. It is a serious threat when it comes to safety-critical applications. Therefore, it is important to generate strong adversarial samples and study the DNN behavior against these adversarial samples. By considering the DNN behavior, the adversarial robustness of DNNs can be improved and more effective defenses can be constructed.
Most attacks and defenses focus on 2D images \cite{goodfellow2015explaining,carlini2017towards,madry2019deep} They are still in the early stages on 3D data \cite{liu2019extending,zhou2019dup,yang2021adversarial}. 
Paying attention to 3D data is interesting, because the world around us is a combination of 3D objects. In addition, 3D data has many applications in robotics, augmented reality, autopilot, and automatic driving. Thanks to the presence of 3D sensors, such as various types of 3D scanners, LiDARs, and RGB-D cameras (such as Kinect, RealSense, and Apple depth cameras), it is easier to capture 3D data. 
This paper proposes an untargeted 3D adversarial point cloud attack against point cloud classifiers; namely, PointNet, PointNet++, and DGCNN. The proposed attack, adds a few points while applying hard boundary constraints on the number of added points and on the point perturbation norms. By controlling the step-size, the generated adversarial sample yields to escape local optima and find the most appropriate attack.
The rest of this paper is organized as follows. In Section 2, the related work is reviewed. The proposed 3D adversarial attack is introduced in Section 3. Experimental results are discussed in Section 4. Finally, Section 5 concludes the paper.
In summary, the contributions of this work include:
\begin{itemize}
\item Proposing an adversarial attack method to perform effective attacks while preserving the point cloud appearance by applying two hard boundary constraints on the number of modified points and on the point perturbation norms.
\item Proposing a learning rate scheduling algorithm to improve other existing methods in this setting.
\item Managing to generate highly successful attacks with a small number of steps to perform fast and subtle attacks.


\end{itemize}

\section{Related Work}

\subsection{Deep Learning on 3D Data}
There are three strategies for 3D object classification including volume-based~\cite{wu20153d, maturana2015voxnet}, multi-view-based~\cite{su2015multi, yang2019learning}, and point cloud-based~\cite{qi2017pointnet,qi2017pointnet1,wang2019dynamic}. This research focuses on point cloud-based models. 
As a pioneering work, the PointNet~\cite{qi2017pointnet} can directly feed point clouds as its input. It achieves the features of each point independently and then aggregates them by max-pooling. It then extracts global features for 3D point cloud classification and segmentation tasks. An update of this work is the PointNet++. It improves the feature extraction through combined features from multiple scales in order to add locality to the PointNet. More recent work apply convolutions on neighborhood points to aggregate more local context~\cite{thomas2019kpconv, hua2018pointwise, wu2019pointconv, li2018pointcnn, wang2019dynamic}. For instance, DGCNN~\cite{wang2019dynamic} processes neighborhood points by applying EdgeConv to better capture local geometric structures of points and therefore achieves superior classification results. This paper uses the PointNet, PointNet++, and DGCNN architectures to evaluate the proposed attack in the case of 3D point cloud classification task.

\subsection{Adversarial Point Clouds}
Various studies have focused on adversarial attack on 3D point cloud classification. The adversarial attacks can be categorized into band-limited and unrestricted adversarial perturbations. 

Band-limited approaches, apply the limitation on the perturbations in generated adversarial samples while preserving the point cloud appearance visually. Typical perturbation measurements include L2 norm, Chamfer distance, and Hausdorff distance. The band-limited attacks are divided into point addition, point shifting, and point dropping attacks. In terms of point addition, Xiang et al. \cite{xiang2019generating} proposed three different targeted attacks by adding several point clusters, tiny objects, or extra points. These attacks optimize a Carlini \& Wagner (C\&W) function \cite{carlini2017towards} and constraint point perturbation norm to push the added points towards the object surface. 
Yang et al. \cite{yang2021adversarial} add a few points to the original point cloud based on the Fast Gradient Sign Method (FGSM) attack \cite{goodfellow2015explaining}. They restrict both the number of added points and the point perturbation norm to generate imperceptible targeted adversarial samples. 
Liu et al. \cite{liu2019adversarial} add new points (sticks or line segments) into the original point cloud, where the sticks must arise from the object's surface. The position of each stick onto the object's surface and the number of points across the line segments are limited.
In addition to generating point clouds by adding points into an original point cloud, both Zheng et al. \cite{zheng2019pointcloud} and Matthew et al. \cite{wicker2019robustness} iteratively drop points from the original point cloud to deceive the classifier. Also, Xiang et al. \cite{xiang2019generating}, Liu et al. \cite{liu2019adversarial, liu2019extending}, Yang et al. \cite{yang2021adversarial}, Tzungyu et al.  \cite{tsai2020robust}, Hamdi et al. \cite{hamdi2020advpc}, and Chengcheng et al. \cite{ma2020efficient} all propose adversarial attack based on point shifting methods. Most of those attacks extend 2D adversarial attacks \cite{carlini2017towards, goodfellow2015explaining, madry2019deep}.

Another line of attacks focuses on unrestricted attacks, which are not limited to any distance criteria. These unlimited attacks do not necessarily look the same as the original point clouds. In other words, it is sufficient that the adversarial sample stays legitimate for the human eye but deceives the classifier. Applying isometry transformation on point cloud \cite{zhao2020isometry} and using trained Generative Adversarial Network (GAN) \cite{zhou2020lg} to generate adversarial sample are some research that has been proposed to unrestricted attacks. Since unlimited attacks do not visually preserve the point cloud appearance, they are not discussed in this paper.

There are typical adversarial defense methods including Statistical Outlier Removal (SOR) \cite{zhou2019dup} and saliency map removal \cite{liu2019extending}, which discard outlier and saliency points, respectively. Also, Zhou et al. \cite{zhou2019dup} propose a denoiser and upsampler network (DUP-Net) structure as defenses for the 3D classification task.

This paper proposes an untargeted attack by a point addition method that imposes hard boundary constraints on the number of added points and on the point perturbation norms. To the best of our knowledge, all previous attacks train with a fixed learning rate until the objective function stagnates, but the proposed attack can escape local optima by controlling the learning rate. Furthermore, by imposing constraints on point perturbation norms and the number of added points it can find the most appearance preserving attacks.

\begin{figure*}
\begin{center}
\includegraphics[width=0.9\linewidth]{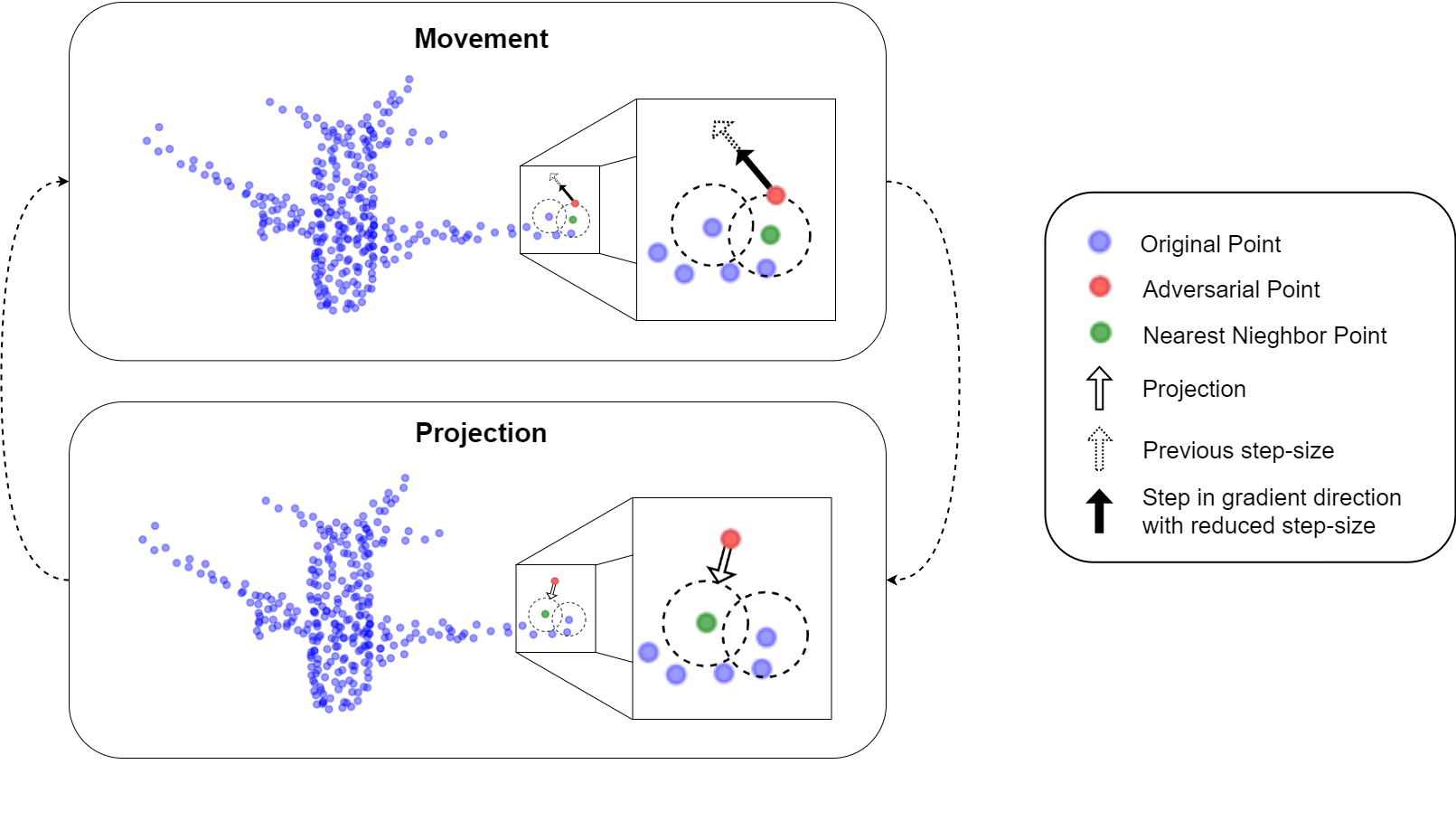}
\end{center}
\caption{Demonstration of proposed Variable Step-Size Attack (VSA) method. At first, adversarial points take a step with reduced step-size in the gradient direction. Then, points that fall out of the original points' boundaries are projected back. This process is repeated at each step.}
\label{fig:method}
\end{figure*}

\section{Proposed Method}
\subsection{Problem Formulation}
The objective is to generate an adversarial sample based on the original one, such that it deceives the model while retaining its appearance. Let $X = \{p_1, p_2, ..., p_k\}$ be the original point cloud, where $p_i \in \mathbb{R}^3$ represents the coordinates of the $i$th point. The adversarial sample $X^* = X \cup \delta_X$ is generated by adding point set $\delta_X$ to the original point cloud. The points in $\delta_X$ are denoted as adversarial points and the points in $X$ are denoted as original points. To deceive the model, the proposed method needs to maximize the model's classification objective function $F(X)$. On the other hand, two hard boundary constraints are applied on the optimization problem to preserve the point cloud's appearance. More concretely, the proposed method is designed to solve the following optimization problem
\begin{equation} \label{eq: problem}
    \max_{\delta_X} F(X^*)\ s.t.\ dist(X^*, X) \leq \epsilon,\ |\delta_X| \leq n
\end{equation}
where $\epsilon$ and $n$ are hard boundaries on the perturbation norm and the number of added points, respectively. The function $dist$ measures the dissimilarity between the adversarial sample and the original one. This function is chosen to be the $\mbox{Hausdorff}$ distance defined by
\begin{equation}
\begin{gathered}
    D_H(X, Y) = \max\{\max_{x_i\in X}\{\min_{y_j\in Y} {||x_i - y_j||}_2\},
    \ \max_{y_j\in Y}\{\min_{x_i\in X} {||y_j - x_i||}_2\}\}
\end{gathered}
\end{equation}
where $x_i$ and $y_j$ are points from point clouds $X$ and $Y$, respectively. This function limits each point from $\delta_X$ to be in the $\epsilon$ neighborhood of its closest neighbor from $X$. In other words, all the added points must be close to the surface of the original point cloud.
    
Although the applied constraints preserve the adversarial sample's appearance, they make the search for the optimal $\delta_X$ much more difficult, especially in a first-order method. The $\mbox{Hausdorff}$ distance constraint limits the movement of the added points which might lead the optimization process to get stuck in bad local maxima. Moreover, due to the low number of added points which leads to low search space dimension, the optimization problem landscape is very non-concave \cite{rottran}. The proposed methods in the following sections overcome these problems to produce better results using first-order methods.

\subsection{Variable Step-Size Attack}
The main proposed method uses the Projected Gradient Descent (PGD) \cite{madry2019deep} algorithm with high step-size at the beginning of the algorithm, to solve Equation \ref{eq: problem}. High step-size lets the added points to explore the whole surface of the point cloud efficiently, while giving them the possibility to escape from local maxima. To converge to the desired result, the step-size is gradually reduced throughout the algorithm.

The algorithm is summarized in Algorithm \ref{alg:vsa}. Suppose that $P$ is the original point cloud. First, $\delta_P$ is initialized using the points of $P$ with the highest gradient norms. These points have the most effect on the classification objective function and are thus a good initialization. At each step, the points in $\delta_P$ take a step in the gradient direction, and in a random direction if their respective gradient norm is zero. At the end of each step, each point in $\delta_P$ is projected into the $\epsilon^*$ neighborhood of its respective nearest neighbor from $P$. The step-size is reduced at each step to ensure the algorithm's convergence. For further insights, the algorithm is demonstrated in Figure \ref{fig:method}.

\begin{algorithm}[tb]
\caption{Variable Step-Size Attack}
\label{alg:vsa}
\textbf{Input}: Objective function $F$, point cloud $P = \{p_1, p_2, ..., p_k\}$\\
\textbf{Parameter}: $\mbox{Hausdorff}$ distance boundary $\epsilon$, initial step-size $\alpha_{init}$, final step-size $\alpha_{final}$, number of steps $\mbox{M}$, number of points $n$ \\
\textbf{Output}: Adversarial sample $P^*$
\begin{algorithmic}[1] 
\STATE $\epsilon^* = \epsilon$
\STATE Initialize $\delta_P$ using the $n$ points with the highest ${||\triangledown_{p_i} F(P)||}_2$.
\FOR{$i=1$ to $\mbox{M}$} 
\STATE Let $\alpha = \alpha_{init} + \frac{\alpha_{final} - \alpha_{init}}{\mbox{M}}$.
\STATE Let $P_0$ and $P_1$ be subsets of $\delta_P$ with ${||\triangledown_{\delta_P} F(P \cup \delta_P)||}_2$ equal to zero and non-zero respectively.
\STATE $P_1 = P_1 +\alpha .  \frac{\triangledown_{P_{1}} F(P \cup \delta_P)}{{||\triangledown_{P_{1}} F(P \cup \delta_P)||}_2}$
\STATE $P_0 = P_0 + \alpha . v$ where $v$ is a random unit vector.
\STATE $\delta_P = P_0 \cup P_1$
\FOR{$p^*_i \in \delta_p$}
\IF{$D_H(p^*_i, P) > \epsilon^*$}
\STATE $p^*_i = \mbox{NN}(p^*_i) + \epsilon^* . \frac{p^*_i - \mbox{NN}(p^*_i)}{{||p^*_i - \mbox{NN}(p^*_i)||}_2}$ where $\mbox{NN}$ is the nearest neighbor in $P$
\ENDIF
\ENDFOR
\ENDFOR
\STATE \textbf{return} $P^* = P \cup \delta_P$
\end{algorithmic}
\end{algorithm}

\subsection{Variable Boundary Attack}
An alternative way to overcome the problem of local maxima is to vary the $\epsilon^*$ parameter. In this method, the $\epsilon^*$ parameter is set high at the beginning of the algorithm and is reduced to the desired final value $\epsilon$, during the algorithm. By having a relaxed constraint for the first steps of the algorithm, it is easier to find solutions with a high objective function value. By proceeding throughout the algorithm, the solutions found with higher $\epsilon^*$ values serve as good initialization points for lower $\epsilon^*$ values. This finally leads to a better solution for the desired $\epsilon$ value at the end of the algorithm, compared to using the PGD algorithm with a constant $\epsilon^*$ and $\alpha$ parameter.

In this algorithm, compared to VSA, the hyperparameters $\alpha_{init}$ and $\alpha_{final}$ are replaced with initial boundary $\epsilon_{init}$ and step-size $\alpha$. Moreover, instead of $\alpha$, $\epsilon^*$ is updated using $\epsilon^* = \epsilon_{init} + \frac{\epsilon - \epsilon_{init}}{\mbox{M}}$.
\section{Experimental Results}
In this section, the proposed method, which is denoted by VSA, is evaluated and compared with other methods to demonstrate its effectiveness. The experiments are carried out on three state-of-the-art point cloud processing architectures run on two benchmark datasets. The proposed method surpasses other state-of-the-art methods in attacking deep point cloud models when using a limited number of points. Moreover, it is shown that the proposed step-size scheduling algorithm can be adopted by existing methods to achieve higher results. The effectiveness of the proposed method against defense methods is also discussed in this section. Moreover, An ablation study is carried out to compare different
variants of the proposed methods and to explore the effects of different hyperparameters. Finally, the generated samples of methods are visualized and compared in terms of perceptibility of adversarial points.
\subsection{Experimental Setup}
\subsubsection{Baselines}
The state-of-the-art methods in the scope of the discussed problem, which include the Point-Attach Method (PAM) in \cite{yang2021adversarial} and the Adversarial Sticks Method (ASM) in \cite{liu2019adversarial}, are employed. For a fair comparison, the methods are chosen to be point addition methods which put hard boundary constraints on the perturbation norms and on the number of added points. In ASM, the farthest point sampling is avoided and new points are sampled onto the adversarial sticks, for it to be used as a point addition method.
\subsubsection{Datasets and Architectures}
The main experiments are carried out against three popular models; namely PointNet \cite{qi2017pointnet}, PointNet++ \cite{qi2017pointnet1}, and DGCNN \cite{wang2019dynamic}. The benchmark datasets used for these experiments are ModelNet40 and ScanObjectNN. The ModelNet40 dataset is used for 3D CAD model classification. The training split of the dataset with $9,843$ samples is used to train the models and the test split with $2,468$ samples is used to evaluate the attack methods. On the other hand, the ScanObjectNN which is a real-world dataset consisting of indoor 3D objects is divided into a training split with $2,309$ samples and a test split with $581$ samples. The experiments are carried out on ModelNet40 against PointNet, if not mentioned otherwise. In all of the experiments the original point clouds have $k=1,024$ points and are normalized according to \cite{qi2017pointnet}.

\subsubsection{Hyperparameters}
All the hyperparameters are initialized according to this section unless mentioned otherwise. For the VSA method, $\alpha_{init}$ is set to $0.1$ and $\alpha_{final}$ is set to $\min\{\frac{0.5}{n}, \frac{\epsilon}{2}\}$. By this, if $n$ is low, the adversarial points spend more time exploring the surface of the point cloud which benefits them since they only cover a small portion of the surface and might need time to reach the optimal solution. Note that according to the observations, the adversarial points tend to distance from each other when proceeding towards the optimal solution. This is because for the studied models in this paper, it is observed that when two points get too close to each other, one overshadows the other's contribution to the classification objective function. Since the points with the most impact on the classification objective function are chosen as the adversarial points' initialization, they tend to be distanced from each other too. Therefore, if $n$ is high, less time is needed to search the point cloud surface since the points already cover the majority of the point cloud surface, which makes them more probable to be close to the optimal solution.

For the Variable Boundary Attack (VBA) method, $\epsilon_{init}$ is set to $2\epsilon$ and $\alpha$ is set to $\min\{\frac{1}{n}, \epsilon\}$. The $\epsilon_{init}$ should be low enough to propose a solution similar enough to the optimal one at the final $\epsilon$, and it should be high enough to solve the problem of local maxima to a certain extent. This makes $2\epsilon$ an appropriate initialization for $\epsilon_{init}$. For $\alpha$, an initialization method similar to that of $\alpha_{final}$ of VSA is chosen due to the reasons discussed in the previous paragraph. The number of steps $M$ is set to $500$ for both methods. 

\subsection{Obtained Results}

All the attack success rates are reported in percentage. The attack success rates against the models trained on ModelNet40 and ScanObjectNN are reported in Tables \ref{table:main} and \ref{table:scan}, respectively. The reported results are against PointNet, PointNet++, and DGCNN in Table \ref{table:main} and against PointNet in Table \ref{table:scan}. The experiments were repeated for different pairs of constraint boundaries $(\epsilon, n)$. Note that the nearest neighbor distance mean for the points in original point clouds (after normalization) is around $0.05$. As such, $\epsilon \in \{0.05, 0.1\}$ makes the adversarial points stay near the point cloud surface. As reported in these tables, the proposed method outperforms other state-of-the-art methods by a large margin. It can be seen that PointNet and PointNet++ are very vulnerable against the proposed method. They almost misclassify every given sample when attacked by adding less than $10\%$ of the points, near the point cloud surface. In contrast, DGCNN performs much better against attack methods and is more challenging. Despite this, the proposed method manages to deceive this model $60\%$ of times. Moreover, as shown in Table 2, the proposed method manages to generate subtle adversarial samples with high accuracy on a real-world dataset. This shows the effectiveness of the proposed method in real-world settings.
\begin{table*}[t]
\caption{Attack success rates (in percentage) on ModelNet40 against PointNet, PointNet++, and DGCNN. Success rates are reported for every pair of constraint boundaries $(\epsilon,\ n)$.}
\centering

\label{table:main}
\begin{tabular*}{\textwidth}{c @{\extracolsep{\fill}} c @{\extracolsep{\fill}} c @{\extracolsep{\fill}} c @{\extracolsep{\fill}} c @{\extracolsep{\fill}} c}
 \noalign{\smallskip}
\hline
 \noalign{\smallskip}
Model & Method & (0.05, 25) & (0.05, 100) & (0.1, 25) & (0.1, 100) \\ 
\noalign{\smallskip}
\hline
\noalign{\smallskip}
\multicolumn{1}{c|}{\multirow{3}{*}{PointNet}}                 & PAM & 2.5 & 11.3 & 8.5 & 36.9 \\
\multicolumn{1}{c|}{}                  & ASM & 6.7 & 17.8 & 17.1 & 33.9 \\
\multicolumn{1}{c|}{}                  & VSA & \textbf{36.5} & \textbf{77.6} & \textbf{88.5} & \textbf{99.5} \\ 
\noalign{\smallskip}
\hline
\noalign{\smallskip}
\multicolumn{1}{c|}{\multirow{3}{*}{PointNet++}}              & PAM & 3.2 & 10.7 & 7.8 & 22.6 \\
\multicolumn{1}{c|}{}                  & ASM & 1.9 & 5.5 & 9.9 & 17.8 \\
\multicolumn{1}{c|}{}                  & VSA & \textbf{21.0} & \textbf{52.5} & \textbf{56.5} & \textbf{97.0} \\ 
\noalign{\smallskip}
\hline
\noalign{\smallskip}
\multicolumn{1}{c|}{\multirow{3}{*}{DGCNN}}        & PAM & 3.4 & 8.0 & 5.1 & 15.2 \\
\multicolumn{1}{c|}{}                  & ASM & 2.7 & 3.5 & 2.8 & 4.8 \\
\multicolumn{1}{c|}{}                  & VSA & \textbf{12.6} & \textbf{29.3} & \textbf{28.7} & \textbf{60.0} \\ 
\noalign{\smallskip}
\hline
\end{tabular*}
\end{table*}

\begin{table*}
\caption{Attack success rates (in percentage) on ScanObjectNN against PointNet. Success rates are reported for every pair of constraint boundaries $(\epsilon,\ n)$.}
\centering
\label{table:scan}
\begin{tabular*}{\textwidth}{c @{\extracolsep{\fill}} c @{\extracolsep{\fill}} c @{\extracolsep{\fill}} c @{\extracolsep{\fill}} c}
 \noalign{\smallskip}
\hline
 \noalign{\smallskip}
Method & (0.05, 25) & (0.05, 100) & (0.1, 25) & (0.1, 100) \\ 
\noalign{\smallskip}
\hline
\noalign{\smallskip}
PAM & 9.1 & 28.8 & 62.3 & 91.1 \\
ASM & 32.8 & 46.7 & 59.1 & 72.5 \\
VSA & \textbf{59.0} & \textbf{84.7} & \textbf{95.1} & \textbf{99.4} \\ 
\noalign{\smallskip}
\hline
\end{tabular*}
\end{table*}
\subsubsection{Adoption by Existing Methods}
Due to the effectiveness of the proposed method, it can be adopted by other methods as well. To assess its effect, the step-size scheduling algorithm was used on PAM and ASM. For ASM, the learning rate was initialized with $2$ instead of $0.1$ and was reduced to $0.01$ throughout the algorithm. That method is denoted as ASM+. For PAM, the step-size is initialized with $\frac{\epsilon}{2}$ and is divided by $2$ at the end of each step. This version of PAM is denoted as PAM+.

The attack success rates of these methods and their enhanced versions are shown in Table \ref{table:transfer}. The results are reported for different pairs of constraint boundaries $(\epsilon, n)$. As shown in this table, the proposed step-size scheduling algorithm is able to improve the existing algorithms for every pair of constraint boundaries. The improvements are more significant for $\epsilon=0.1$, especially for the ASM  algorithm where an improvement of $14.5\%$ is made with $(0.1, 100)$ as parameters. The difference in improvement between ASM and PAM is most likely due to the usage of projection in ASM's algorithm, which leads to a more effective search when paired with the step-size scheduling algorithm. Overall, this shows the impact of the proposed search strategy on other algorithms, in the scope of this problem.

\subsubsection{Robustness Against Defense Methods}
In this section, the proposed method is evaluated against statistical outlier removal (SOR) defense and salient point removal (SPR) defense. For the outlier removal defense, the $10$ nearest neighbor average distance is calculated for each point. The points that have an average distance of greater than one standard deviation from the mean of this statistic are removed. For the salient point removal, the $200$ points with the highest saliencies are removed.

The attack success rates against these two defense methods are reported in Table \ref{table:defense}. For each number of points $n\in \{25, 100\}$, different values for $\epsilon\leq 0.1$ were tested and the best result was reported. The SOR defense is more challenging when the number of points is low. This comes from the fact that the mean of the statistic calculated for the original point cloud does not change drastically when the adversarial points are added. Despite its challenges, the proposed method manages to evade the defense by adding the adversarial points very close to the point cloud surface with $\epsilon\in \{0.025, 0.05\}$, while having a high classification objective function. This is why it outperforms other methods against SOR. However, it is not as effective as ASM when it comes to SPR. This is because in ASM, when the salient points which are usually on the head of the sticks are removed, there are other close adversarial points that will replace the head of the sticks and make a successful attack.

\begin{table*}
\caption{Attack success rates (in percentage) of existing methods and their step-size scheduling enhanced counterparts. Success rates are reported for different pairs of constraint boundaries $(\epsilon, n)$ on ModelNet40 against PointNet.}
\centering
\label{table:transfer}
\begin{tabular*}{\textwidth}{c @{\extracolsep{\fill}} c @{\extracolsep{\fill}} c @{\extracolsep{\fill}} c @{\extracolsep{\fill}} c}
 \noalign{\smallskip}
\hline
 \noalign{\smallskip}
Method & (0.05, 25) & (0.05, 100) & (0.1, 25) & (0.1, 100) \\ 
\noalign{\smallskip}
\hline
\noalign{\smallskip}
PAM & 2.5 & 11.3 & 8.5 & 36.9  \\
PAM+ & \textbf{3.7} & \textbf{12.8} & \textbf{10.7} & \textbf{42.7} \\
\noalign{\smallskip}
\hline
\noalign{\smallskip}
ASM & 6.7 & 17.8 & 17.1 & 33.9\\
ASM+ & \textbf{10.5} & \textbf{24.4} & \textbf{30.8} & \textbf{48.4} \\ 
\noalign{\smallskip}
\hline
\end{tabular*}
\end{table*}

\begin{table*}
\caption{Attack success rates (in percentage) against outlier removal defense and salient point removal defense. Success rates are reported for every pair of constraint boundaries $(\epsilon, n)$ on ModelNet40 against PointNet. The best success rate for each number of points $n$ is chosen for comparison.}
\centering
\label{table:defense}
\begin{tabular*}{\textwidth}{c @{\extracolsep{\fill}} c @{\extracolsep{\fill}} c @{\extracolsep{\fill}} c @{\extracolsep{\fill}} c
@{\extracolsep{\fill}} c @{\extracolsep{\fill}} c @{\extracolsep{\fill}} c @{\extracolsep{\fill}} c
@{\extracolsep{\fill}} c}
 \noalign{\smallskip}
\hline
 \noalign{\smallskip}
Defense &  Method & (0.025, 25) & (0.05, 25) & (0.1, 25) & \multicolumn{1}{c|}{25} & (0.025, 100) & (0.05, 100) & (0.1, 100) & 100 \\ 
\noalign{\smallskip}
\hline
\noalign{\smallskip}
\multirow{3}{*}{SOR} 
 & PAM & 4.4 & 5.5 & 4.0 & \multicolumn{1}{c|}{5.5} & 6.8 & 7.7 & 3.6 & 7.7 \\
 & ASM & 5.2 & 6.8 & 6.5 & \multicolumn{1}{c|}{6.8} & 2.9 & 11.9 & 18.1 & 18.1  \\
 & VSA & 13.0 & 13.0 & 2.8 & \multicolumn{1}{c|}{\textbf{13.0}} & 28.0 & 31.8 & 9.5 & \textbf{31.8} \\
\noalign{\smallskip}
\hline
\noalign{\smallskip}
\multirow{3}{*}{SPR} 
 & PAM & 1.9 & 2.5 & 1.6 & \multicolumn{1}{c|}{2.5} & 1.6 & 2.3 & 2.3 & 2.3\\
 & ASM & 2.3 & 3.7 & 7.1 & \multicolumn{1}{c|}{\textbf{7.1}} & 5.4 & 6.5 & 13.7 & \textbf{13.7}  \\
 & VSA &4.2 & 1.1 & 4.6 &  \multicolumn{1}{c|}{4.6} & 7.1 & 8.2 & 8.5 & 8.5\\
\noalign{\smallskip}
\hline
\end{tabular*}
\end{table*}

\begin{table*}
\caption{Attack success rates (in percentage) of the proposed methods and their variants. Success rates are reported for every pair of constraint  boundaries $(\epsilon, n)$ on ModelNet40 against PointNet.}
\centering
\label{table:variants}
\begin{tabular*}{\textwidth}{c @{\extracolsep{\fill}} c @{\extracolsep{\fill}} c @{\extracolsep{\fill}} c}
 \noalign{\smallskip}
\hline
 \noalign{\smallskip}
Method & (0.05, 25) & (0.05, 100) & (0.05, 400)\\ 
\noalign{\smallskip}
\hline
\noalign{\smallskip}
 PGD & 32.6 & 67.0 & 86.4  \\
VBA & 34.2 & 69.8 & 87.5 \\
$\mbox{VSA}^*$ & 30.5 & 70.1 & 90.6\\
VSA & 36.5 & \textbf{77.6} & 93.9 \\
VBA + VSA & \textbf{37.4} & 76.4 & \textbf{94.5} \\ 
\noalign{\smallskip}
\hline
\end{tabular*}
\end{table*}

\begin{table*}
\caption{Attack success rates (in percentage) for different number of steps values $M$ and different number of points values $n$ with $\epsilon=0.05$. Success rates are reported on ModelNet40 against PointNet.}
\centering
\label{table:parameters}
\begin{tabular*}{\textwidth}{c  c @{\extracolsep{\fill}} c @{\extracolsep{\fill}} c @{\extracolsep{\fill}} c @{\extracolsep{\fill}} c @{\extracolsep{\fill}} c}
 \noalign{\smallskip}
\hline
 \noalign{\smallskip}
 \multirow{2}{*}{n} & \multicolumn{6}{|c}{M} \\
 \noalign{\smallskip}
 \cline{2-7}
 \noalign{\smallskip}
 & \multicolumn{1}{|c}{25} & 50 & 100 & 300 & 500 & 800 \\ 
\noalign{\smallskip}
\hline
\noalign{\smallskip}
 \multicolumn{1}{c|}{25} & 28.7 & 31.6 & 33.4 & 36.4 & 36.5 & 38.6 \\
\multicolumn{1}{c|}{50} & 49.2 & 52.1 & 54.6 & 59.0 & 59.5 & 60.0 \\
\multicolumn{1}{c|}{100} & 67.0 & 71.1 & 73.9 & 76.4 & 77.6 & 77.6  \\
 \multicolumn{1}{c|}{200} &  80.0 & 83.3 & 85.1 & 88.3  & 88.4 & 88.4 \\ 
  \multicolumn{1}{c|}{400} & 87.3 & 89.6 & 91.1 & 92.8 & 93.9 & 93.8 \\ 
\noalign{\smallskip}
\hline
\end{tabular*}
\end{table*}
\subsection{Ablation}
In this section, different variants of the proposed methods are compared to each other and different aspects of the proposed method are explored and evaluated. The simplest variant of the proposed method is the PGD algorithm, where the step-size $\alpha$ is constant compared to VSA. In this method, $\alpha$ is initialized according to VBA. A more complicated version of the proposed methods is a method comprising both of their ideas, denoted as VBA + VSA. In this method, the $\epsilon^*$ variable is scheduled according to VBA and $\alpha$ is scheduled according to VSA. Moreover, a variant of VSA where $a_{init}$ is not set high (it is set to $0.025$, $0.02$, and $0.01$ for $n$ equal to $25$, $100$, and $400$, respectively) is also considered. $\alpha_{final}$ is set to $0.01$ for $n=25$ in that method. The attack success rates for different variants of the proposed methods are reported in Table \ref{table:variants}. The experiments were carried out for different pairs of constraint boundaries. As shown in this table, VBA outperforms PGD by solving the problem of local maxima to a certain extent. Moreover, the methods improved by the proposed step-size scheduling algorithm outperform the other methods including $\mbox{VSA}^*$ to a large extent. It can be seen that $\alpha_{init}$ needs to be set high for VSA to work effectively, especially when $n$ is set low. The improvement made by the step-size scheduling algorithm is due to a  number of factors, like escaping improper local maxima, being able to explore the point cloud surface better, and converging to the desired result at the end of the algorithms. Between the algorithms that employ the proposed step-size scheduling, VBA + VSA performs slightly better. However, it is observed that VSA performs slightly better in Table \ref{table:parameters} experiments when compared to VBA + VSA. Due to its slightly better performance and its simplicity, the VSA was chosen as the main proposed method.

Table \ref{table:parameters} contains attack success rates for different values of $M$ and $n$ with $\epsilon=0.05$. It can be seen that the proposed method performs very effectively, even when the number of steps is as low as $25$. This makes the proposed method ideal for fast and subtle attacks. Note that a higher number of steps lets the adversarial points better explore the point cloud surface, though its impact slowly starts to decrease as the number of steps increases. Moreover, when the number of points is higher, the algorithm gets less affected by decreasing the number of steps, which is due to the effective initialization discussed in Section 4.1.3. To explore whether the local maxima problem is tackled or the improvements of VSA are merely due to the high number of steps and better exploration of the point cloud surface, the results in Table \ref{table:parameters} are compared with Table \ref{table:variants}. Consider $n\in\{25, 100\}$ where the local maxima problem is worse. By comparing the success rates of $(M=100, n=25)$ and $(M=100, n=100)$ in Table \ref{table:parameters} to $(\epsilon=0.05, n=25)$ and $(\epsilon=0.05, n=100)$ of the $\mbox{VSA}^*$ method in Table \ref{table:variants} respectively, it can be seen that the respective success rates are higher. However, the respective traversed distance per points are lower. This shows that this improvement is due to solving the local maxima problem.

\subsection{Visualization}
\begin{figure*}
\begin{center}
\includegraphics[width=0.92\linewidth]{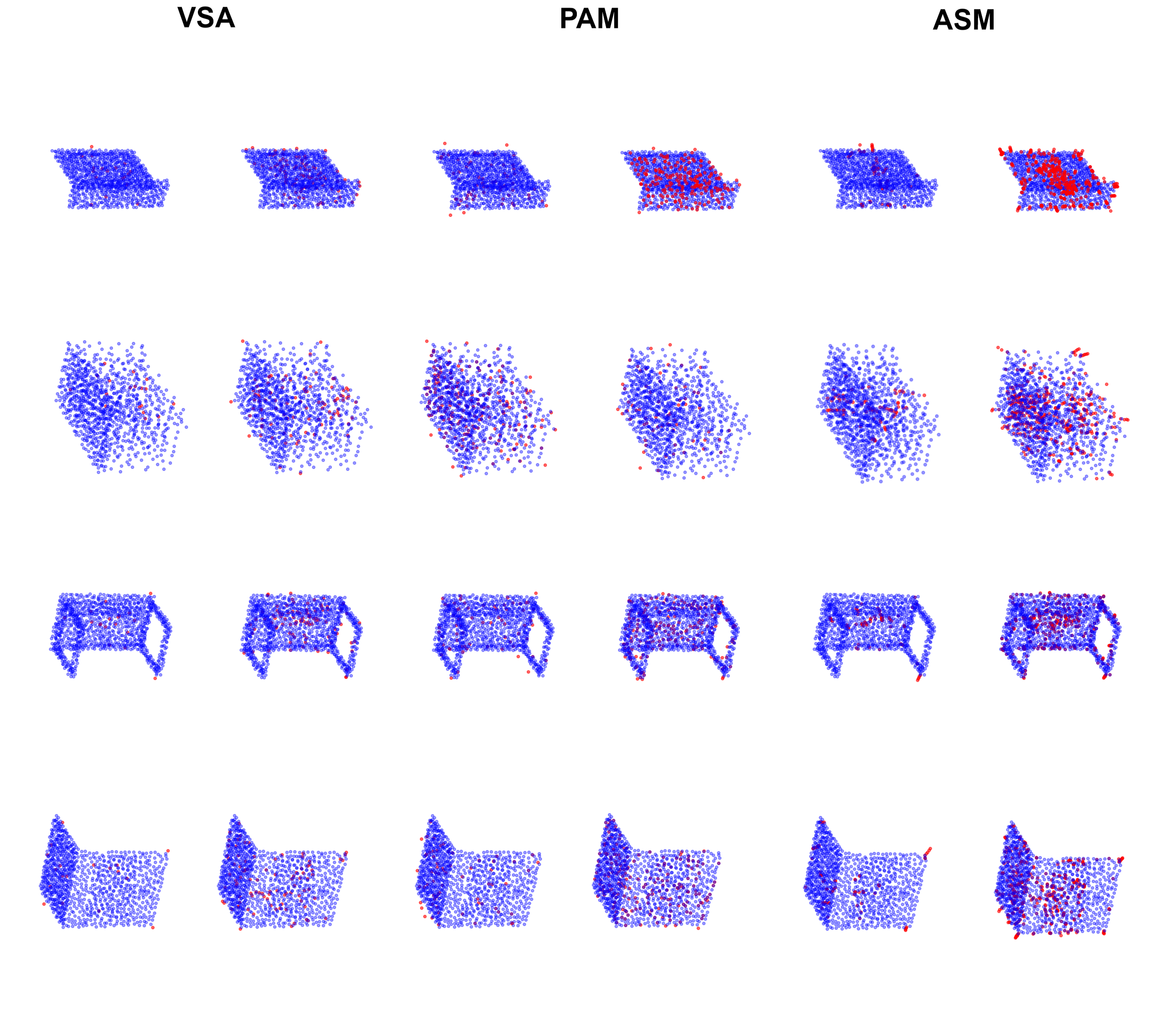}
\end{center}
\caption{Adversarial Samples generated by each method with $(\epsilon, n)$ set to reach $25\%$ attack success rate. For each method, first column represents low $n$ and high $\epsilon$ and second column represents high $n$ and low $\epsilon$. Red points represent adversarial points and Blue points represent original points.}
\label{fig:vis}
\end{figure*}

In this section, the perceptibility of the generated adversarial sample is explored. For this, the attack success rate is fixed on $25\%$ and $(\epsilon\in[0.025, 0.1], n\in[25, 400])$ are chosen for each of the methods to reach this success rate threshold. The generated samples are shown in Figure \ref{fig:vis}. Since there is a trade-off between $n$ and $\epsilon$ to increase success rate, the first column for each method contains samples with low $n$ and high $\epsilon$ and the second column contains samples with high $n$ and low $\epsilon$.

As shown in Figure \ref{fig:vis}, the adversarial points are far less perceptible in the proposed method compared to existing methods. Moreover, it is more outlier free due to its ability to perform successful attacks with low $\epsilon$.

\section{Conclusion}
This paper proposed a new attack method to generate effective adversarial attacks by adding a limited number of points to the point cloud surface. The method introduced an effective step-size scheduling algorithm to overcome the local maxima problem and to explore the point cloud surface efficiently. The results showed that in addition to achieving state-of-the-art results, it can be adopted by other existing methods to improve their results. It also showed that the proposed method performs well against the SOR defense which is more challenging when the number of added points is low. Overall, this shows that 3D deep learning models are vulnerable to subtle yet effective attacks. By these observations, future work could investigate different step-size scheduling algorithms and their effects on the performance of first order attacks. 


\bibliography{iclr2022_conference}
\bibliographystyle{iclr2022_conference}


\end{document}